%% file: main.tex
\documentclass[10pt,twocolumn,letterpaper]{article}

\usepackage{cvpr}              

\definecolor{cvprblue}{rgb}{0.21,0.49,0.74}
\usepackage[pagebackref,breaklinks,colorlinks,allcolors=cvprblue]{hyperref}
\usepackage{natbib}
\usepackage{colortbl}
\definecolor{lightgreen}{rgb}{0.84, 1.0, 0.84}
\usepackage{multirow}

\usepackage{pifont} 
\usepackage{booktabs} 
\usepackage{adjustbox}
\usepackage{arydshln}
\usepackage{tabularx}

\usepackage{pifont} 

\usepackage{tcolorbox}
\tcbuselibrary{breakable,skins} 

\newtcbox{\hlblue}{on line, arc=1pt, colback=blue!12,  boxrule=0pt, boxsep=1pt,
  left=2pt,right=2pt, top=1pt,bottom=1pt}
\newtcbox{\hlyellow}{on line, arc=1pt, colback=yellow!12, boxrule=0pt, boxsep=1pt,
  left=2pt,right=2pt, top=1pt,bottom=1pt}
\newtcbox{\hlgreen}{on line, arc=1pt, colback=green!12, boxrule=0pt, boxsep=1pt,
  left=2pt,right=2pt, top=1pt,bottom=1pt}
\newtcbox{\hlorange}{on line, arc=1pt, colback=orange!12, boxrule=0pt, boxsep=1pt,
  left=2pt,right=2pt, top=1pt,bottom=1pt}
\newtcbox{\hlpurple}{on line, arc=1pt, colback=purple!12, boxrule=0pt, boxsep=1pt,
  left=2pt,right=2pt, top=1pt,bottom=1pt}
\newtcolorbox{promptbox}{
  breakable,                  
  width=\linewidth,
  colback=gray!10,            
  colframe=gray!30,           
  arc=2mm,                    
  boxrule=0.3pt,              
  left=6pt,right=6pt,top=6pt,bottom=6pt
}

\def\paperID{15295} 
\def\confName{CVPR}
\def\confYear{2025}

\title{Insight-A: Attribution-aware for Multimodal Misinformation Detection}

\author{
Junjie Wu$^1$ \ \  \  Yumeng Fu$^3$ \ \  \    Chen Gong$^{1,2}$  \ \  \   Guohong Fu$^{1,2}$\thanks{Corresponding author.}\\
$^1$School of Computer Science and Technology, Soochow University\\
$^2$Institute of Artificial Intelligence, Soochow University\\
$^3$School of Computer Science and Technology, Harbin Institute of Technology\\
{\tt\small 20224027010@stu.suda.edu.cn, 24b303004@stu.hit.edu.cn, \{gongchen18, ghfu\}@suda.edu.cn}
}

\begin{document}
\maketitle

\input{0_abstract}    
\input{1_intro}
\input{2_related}
\input{3_methodology}
\input{4_experiment}
\input{5_conclusion}
{
    \small
    \bibliographystyle{ieeenat_fullname}
    \bibliography{main}
}


\end{document}

%% file: 0_abstract.tex
\begin{abstract}
AI-generated content (AIGC) technology has emerged as a prevalent alternative to create multimodal misinformation on social media platforms, posing unprecedented threats to societal safety. However, standard prompting leverages multimodal large language models (MLLMs) to identify the emerging misinformation, which ignores the misinformation attribution. To this end, we present Insight-A, exploring attribution with MLLM insights for detecting multimodal misinformation. Insight-A makes two efforts: I) attribute misinformation to forgery sources, and II) an effective pipeline with hierarchical reasoning that detects distortions across modalities. Specifically, to attribute misinformation to forgery traces based on generation patterns, we devise cross-attribution prompting (CAP) to model the sophisticated correlations between perception and reasoning. Meanwhile, to reduce the subjectivity of human-annotated prompts, automatic attribution-debiased prompting (ADP) is used for task adaptation on MLLMs. Additionally, we design image captioning (IC) to achieve visual details for enhancing cross-modal consistency checking. Extensive experiments demonstrate the superiority of our proposal and provide a new paradigm for multimodal misinformation detection in the era of AIGC.
\end{abstract}

%% file: 1_intro.tex
\section{Introduction}
\label{sec:intro}

The proliferation of AI-generated content (AIGC) technology has lowered the barrier for malicious fabricators to produce multimodal misinformation \cite{NIPS2014_f033ed80,Wang_2025_CVPR}. This has flooded social media platforms, posing threats to societal safety \cite{Zhang_Zhang_Zhou_Huang_Li_2024,Guo_Ma_Zeng_Luo_Zeng_Tang_Zhao_2025}. Traditional methods to the task of multimodal misinformation detection have focused on AI-generated benchmarks, where they leverage supervised machine learning frameworks to identify the authenticity of news captions, news images, or image-text pairs \cite{wang2024mfcbenchbenchmarkingmultimodalfactchecking,huang_etal_2024_miragenews}. This paradigm has now been disrupted by the emergence of both advanced multimodal large language models (MLLMs) and mixed-source multimodal misinformation \cite{Qwen2VL,liu2024mmfakebench}. Effective multimodal misinformation detection is thus crucial to the credibility of multimodal content for the general public.

\input{latex/fig_1}

In the era of AIGC, multimodal misinformation evolves to multiple and random sources, including textual veracity distortion (TVD), visual veracity distortion (VVD), and cross-modal consistency distortion (CCD). While many studies exist for probing surface features to detect mixed-source multimodal misinformation \cite{beigi2025llmsimprovemultimodalfactchecking,ali-etal-2025-detection}, research on the inherent attribution of each distortion remains under-explored. Previous work treats generation patterns as a critical signal to capture the distinction between different distortions, thereby accurately distinguishing multimodal misinformation. Firstly, as illustrated in Figure~\ref{fig:1}, a dedicated detector is used to judge the generation pattern, with its probability incorporated into a human-crafted prompt for the final judgment \cite{Xu_MDAM3}. However, such proposal depends primarily on large-scale datasets annotated with high-quality labels, which are time-consuming and expensive. On the other hand, a low-resource approach is to utilize standard prompting (SP) for directly querying MLLMs in a zero-shot setting \cite{huang_etal_2024_miragenews}. Nonetheless, this direct query method struggles to make correct decisions due to the lack of mimicking human-like reasoning. Hence, it is essential to explore generation patterns with effective reasoning for better understanding the characteristics of distinct distortions in the context of multimodal misinformation detection.

In this paper, our goal is to track attribution for multimodal misinformation detection. However, identifying multimodal misinformation attribution is non-trivial. Firstly, the subjectivity of human-crafted queries around multimodal misinformation attribution is inevitable to encounter language biases, such as rare words or grammatical mistakes. Secondly, the choice of the generation pattern introduces challenges in bridging the sophisticated correlations between perception and reasoning. Finally, the detailed description from news images is essential for compensation, enhancing consistency checking across modalities.

To this end, we present a novel zero-shot framework, dubbed Insight-A, exploring \underline{A}ttribution with MLLM \underline{Insight}s for effective multimodal misinformation detection. Specifically, we first design automatic attribution-debiased prompting (ADP) to eliminate language biases from human-crafted queries. Subsequently, a cross-attribution prompting (CAP) strategy achieves the reasoning paths across different generation patterns, and allocates alignment scores to each generation pattern and its reasoning path for accurate reasoning answers. Moreover, image captioning (IC) is designed to provide visual details, thus integrating clues into decision-making for debunking multimodal misinformation.

We conduct experiments to compare our Insight-A with advancing state-of-the-art (SOTA) methods on the mixed-sourced multimodal misinformation benchmark dataset MMFakeBench. Experimental results illustrate the superiority of Insight-A and further provide an in-depth analysis of different methods for future works in the era of AIGC. 

In summary, our major contributions are as follows:

\begin{itemize}
    \item We present a novel method, dubbed Insight-A, exploring attribution with MLLM insights for detecting multimodal misinformation in a zero-shot setting.
    \item We introduce a cross-attribution prompting method that facilitates the accurate decision-making more heavily from reasoning paths with higher credibility.
    \item Extensive experiments on the MMFakeBench dataset demonstrate the effectiveness and superiority of Insight-A, and our method of identifying distinct veracity distortions related to multimodal misinformation.
\end{itemize}

%% file: latex/fig_1.tex
\begin{figure}[t]
  \includegraphics[width=\columnwidth]{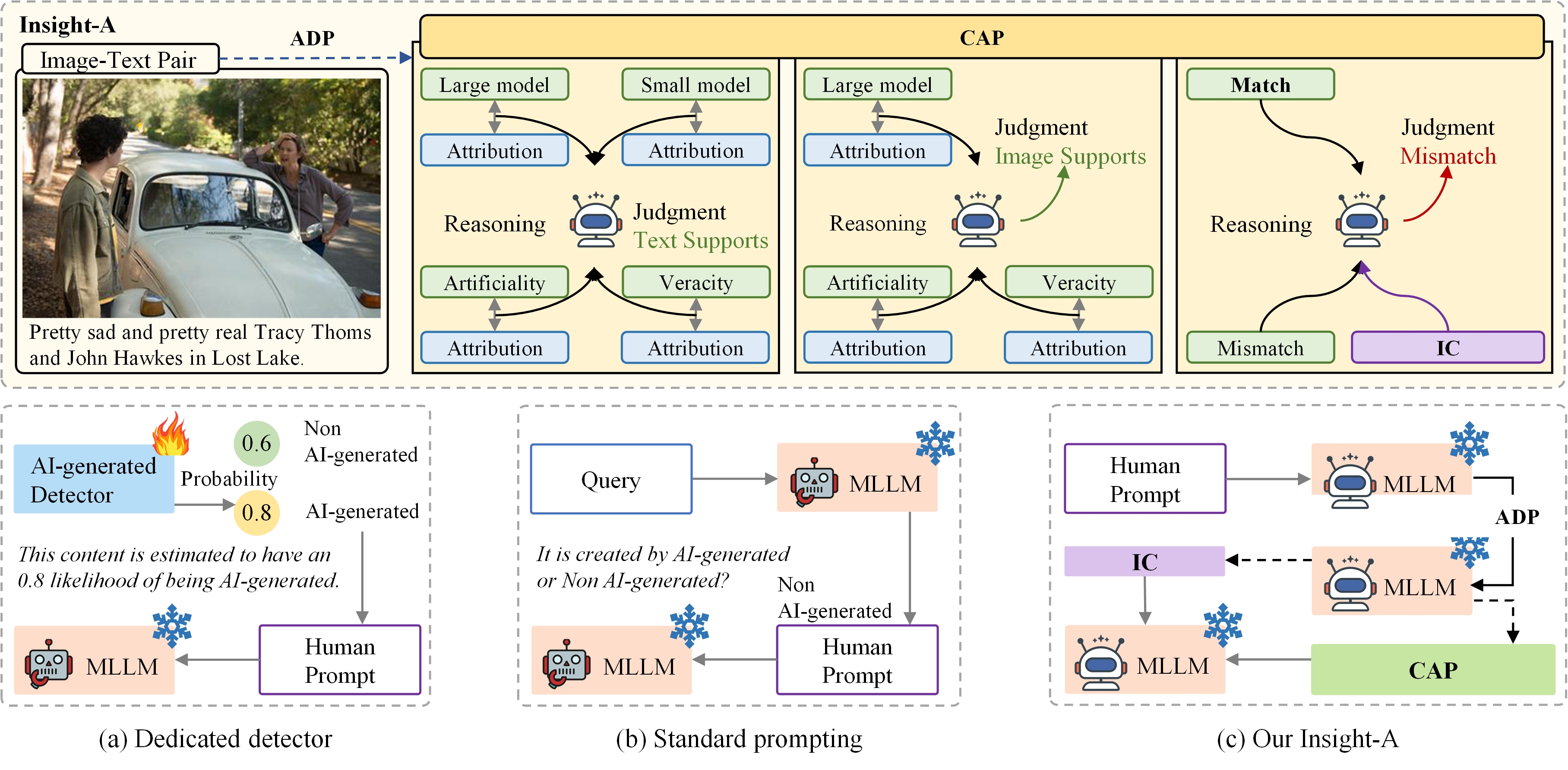}
  \caption{Misinformation detection methods (a) and (b) vs our Insight-A (c). Previous methods either train dedicated detectors, or use standard prompting to combat misinformation. In contrast, our Insight-A attributes multimodal content to generation patterns, thus making accurate decisions.}
  \label{fig:1}
\end{figure}

%% file: 2_related.tex
\section{Related Work}
\label{sec:Related Work}

\subsection{Multimodal Misinformation Detection}

Most of existing works on the task of multimodal misinformation detection use deep models with supervised learning, which can be divided into the following three types: 1) some methods focus on textual veracity in human feedback \cite{Min_2022}, pre-trained language models (PLMs) \cite{zhang_gao_2023_towards}, and style \cite{wang2024mfcbenchbenchmarkingmultimodalfactchecking}. 2) some researchers are dedicated to visual veracity by extracting features based on sentiment \cite{10440475}, background \cite{wang2024mfcbenchbenchmarkingmultimodalfactchecking}, and pre-trained visual models \cite{Liu_dino}. 3) multimodal misinformation strengthens the deception for the general public. Extracting multimodal features, and then fusing them captures the inconsistencies between both news captions and images. Some studies design various attention-based modules for cross-modal fusion \cite{Ying_Hu_Zhou_Qian_Zeng_Ge_2023,chen_etal_2023_causal,Yifang_Moment_GPT}. In this paper, we rely on the reasoning and generation capabilities of MLLMs, thus achieving accurate judgments regarding multimodal misinformation.

\subsection{MLLMs for Multimodal Misinformation Detection}

With the rise and application of large language models (LLMs), numerous works have paved the way for enhancing the perceptual capacities of LLMs regarding visual content \cite{baichuan2023baichuan2,Jiang_2024_CVPR,Wang_2025_CVPR}. This promotes the development of MLLMs in multimodal and visual domains \cite{beigi2025llmsimprovemultimodalfactchecking,Restrepo_2025,Fang_2025_CVPR,Liu_Li_Rao_Gao_Guan_Li_Ma_2025,Yifang_Moment_GPT}. In the realm of multimodal misinformation detection, recent methods leverage diverse series of MLLMs to analyze news without post-training \cite{liu2024mmfakebench}. More importantly, in order to identify mixed-source multimodal misinformation in the era of AIGC, a forerunner uses a hierarchical decomposition paradigm to achieve reasoning answers in a zero-shot setting \cite{liu2024mmfakebench}. Despite their impressive performance, they do not consider multimodal misinformation attribution. In contrast, our work is to explore attribution with MLLM insights on multimodal misinformation.

%% file: 3_methodology.tex
\section{Methodology}
\label{sec:Methodology}

\input{latex/fig_2}


\subsection{Problem Definition}

Multimodal misinformation detection is the task of identifying the authenticity of a given image-text pair. Formally, let $\mathcal{D}=\{(x_i^t,x_i^v,y_i)|(x_i^t,x_i^v)\in\mathcal{X},y_i\in\mathcal{Y}\}_{i=1}^{N}$ represent the set of image-text pairs $\mathcal{X}$ and their labels $\mathcal{Y}$. The symbols $x_i^t$ and $x_i^v$ denote the $i$-th news caption and its attached news image. $N$ refers to the number of all image-text pairs in $\mathcal{D}$. Given a pair of image-text $(x_i^t,x_i^v)$, the goal of a multimodal misinformation detector is to provide a prediction $y\in\{0,1,2,3\}$, where $y=0$ denotes the real item, $y=1$ denotes the textual veracity distortion (TVD), $y=2$ denotes the visual veracity distortion (VVD), and $y=3$ denotes the cross-modal consistency distortion (CCD). The latter three are classified as the fake item.

The task of multimodal misinformation detection is regarded as a multiclass classification issue, where the model $f$ learns a function: $f(\mathcal{X})\to\mathcal{Y}$.

\subsection{Attribution-Debiased Prompting}

Empirical studies have shown that MLLMs exhibit the predicted consistency and instruction-following capabilities \cite{Cai_2025_CVPR,Yao_Yuguang_Pan_Ning_Ye_Zhou_Xie_2025,Yifang_Moment_GPT}. In a zero-shot setting, an effective approach is to use human-crafted prompts, for guiding MLLMs to probe multimodal misinformation detection and attribution. However, this typically includes language biases, such as rare words and grammatical mistakes, due to the subjectivity of human. This constrains MLLMs for task adaptation, thereby reducing the performance of multimodal misinformation detection \cite{Zhang_2024_CVPR}.

To address this limitation, we introduce automatic attribution-debiased prompting (ADP). It leverages a MLLM to simulate human-like behaviors \cite{ye2025generativepsycholexicalapproachconstructing}, thereby generating debiased prompts with the same semantics as manually annotated prompts for task adaptation of the MLLM. Based on the intention, we craft the query to the MLLM for natural language processing as follows:

\begin{promptbox}
\textbf{Query:} \; \dots\ your task is to generate a sentence, please follow the instructions below:\par
1. \hlyellow{Eliminate language biases} including rare words and grammatical mistakes in the Raw Sentence.\par
2. Only return a new prompt while ensuring that it aligns with the \hlgreen{unchanged semantics} of the Raw Sentence.

\medskip
\textbf{Input:} \; \hlblue{[Raw Sentence $\mathcal{S}_{ADP}$]}
\end{promptbox}

Specifically, the ADP directs MLLMs to perform language bias elimination and semantic alignment by analyzing the given query $\mathcal{Q}_{init}$. Formally, the generated sentence can be obtained by:
\begin{equation}
\mathcal{R}_{ADP} = \arg\max P(\mathcal{R} \mid \mathcal{Q}_{init}, S_{ADP}),
\end{equation}
where $\mathcal{R}_{ADP}$ denotes the generated sentence based on the query $\mathcal{Q}_{init}$ and raw sentence $S_{ADP}$. From the results of Table \ref{tab_5}, the ADP can effectively enhance the model to probe multimodal misinformation attribution for improving detection.

\subsection{Cross-Attribution Prompting}

After generating the query from the ADP, as shown in Figure \ref{fig2}, we further design cross-attribution prompting (CAP) to make rigorous and correct decisions more heavily from attribution reasoning with higher credibility. In contrast to illogical repetitive answers from MLLMs \cite{barez_chain_2025}, the CAP integrates generation patterns with hierarchical attribution reasoning. This focuses on capturing the attribution of multimodal misinformation in a high-quality reasoning and high-confidence classification manner. In response to such ideas, the designed CAP conducts the following solution from the perspectives of generation pattern, attribution reasoning, and cross-attribution scoring.

\paragraph{Generation Pattern.} Following the definition of multimodal misinformation in the era of AIGC \cite{huang_etal_2024_miragenews,liu2024mmfakebench}, we investigate generation categories of textual veracity distortion and visual veracity distortion. In textual veracity distortion, forgery captions are created by $\mathcal{P}^T=$ \textit{\{Largemodel, Smallmodel, Artificiality\}}. In visual veracity distortion, forgery images are made from $\mathcal{P}^V=$ \textit{\{Largemodel, Artificiality\}}. In our initial experiments, we have observed the fact that large models struggle to understand these specific words related to generation categories. To improve the MLLM's understanding, basic definitions are essential to $\{\mathcal{P}^T, \mathcal{P}^V\}$ for enhancing MLLM's reasoning beyond words, as shown in Figure \ref{fig2}.

\paragraph{Attribution Reasoning.} Previous studies have shown that, although existing MLLMs have robust reasoning and generation capabilities in extensive domains, they may generate hallucinations for human preferences \cite{li_attributed,weinzierl_tree,Jiang_2024_CVPR}. In contrast to standard prompting for reasoning a certain generation category, we conduct fine-grained MLLM's reasoning for the news towards each potential generation category.

Specifically, depending on one generation category $p^t_i$ from $\mathcal{P}^T$, the language of the ADP is regarded as a query $\mathcal{Q}_{ADP}$, which is introduced into the MLLM for attribution reasoning. The news caption $x_i^t$ is also inserted into the query. Following this, the MLLM yields the response as:
\begin{equation}
\mathcal{R}_{p_i^t} = \arg\max P(\mathcal{R} \mid \mathcal{Q}_{ADP}, x_i^t, p_i^t),
\end{equation}
\noindent where, $\mathcal{R}_{p_i^t}$ denotes the generated reasoning path with multiple steps in the news caption $x_i^t$ and a certain generation category $p_i^t$. Furthermore, the attribution reasoning is performed in other generation categories from $\mathcal{P}^T$. Subsequently, as for the news image $x_i^v$, the MLLM produces the result by:
\begin{equation}
\mathcal{R}_{p_i^v} = \arg\max P(\mathcal{R} \mid \mathcal{Q}_{ADP}, x_i^v, p_i^v),
\end{equation}
\noindent where, $\mathcal{R}_{p_i^v}$ denotes the yielded reasoning path with multiple steps in the news image $x_i^v$ and one generation category $p_i^v$ from the set $\mathcal{P}^V$.

\paragraph{Cross-Attribution Scoring.} After performing attribution reasoning in both news captions and images, we further introduce cross-attribution scoring to assign an alignment score for the attribution of the news. Specifically, the first step is to guide the MLLM to allocate respective scores to each attribution reasoning $R_p$ based on its quality evaluation, denoted as $s^r=\{s^r_1,s^r_2,...,s^r_M\}$. The second step is to query the MLLM for the likelihood of each generation category $\mathcal{P}$ related to the news, denoted as $s^p=\{s^p_1,s^p_2,...,s^p_M\}$. Eventually, we integrate these two scores for consistency across attributions. The whole process is formulated as follows:
\begin{equation}
\mathcal{R}_{CAP} = \arg\max_{i\in M} \;  s_i^r\cdot s_i^p,
\end{equation}
\noindent where $s^r$ and $s^p$ are independent during the process of cross-attribution scoring. This is beneficial to the model for comprehensive evaluation, rather than one-path reasoning.


\subsection{Image Captioning}

By selecting a possible generation category towards both news captions and news images, as shown in Figure \ref{fig2}, we incorporate these clues into the model for judging the authenticity of text or image. However, there may be semantic inconsistencies between them. To enhance cross-modal alignment, we conduct cross-modal consistency checking within the final decision-making. Considering that the news image typically contains visual redundancy, leading to the disturbance in the sense \cite{wu_etal_2022_rap,Yifang_Moment_GPT}.

Inspired by previous work, visual signals in the form of linguistic descriptions can enable the model to effectively comprehend informative contents related to the news image \cite{7346469,Zhang_2025_CVPR}. In response to this characteristic, we perform image captioning (IC) to give the textual content of the news image for improving cross-modal consistency checking. As shown in Figure \ref{fig2}, the descriptive words (e.g., ``snowboarder'' and ``snowy mountain''), which correspond well with the news. Finally, we incorporate this description alongside the news into the MLLM for facilitating informed decision-making:
\begin{equation}
\mathcal{R}_{end} = \arg\max P(\mathcal{R} \mid \mathcal{Q}_{end}, x_i^t, x_i^v, x_i^{v'}),
\end{equation}
where, $\mathcal{Q}_{end}$ and $\mathcal{R}_{end}$ denote the query and the final answer, and $x_i^{v'}$ represents the textual content corresponding to the news image.

%% file: latex/fig_2.tex
\begin{figure*}[t]
\centering
\includegraphics[width=\textwidth]{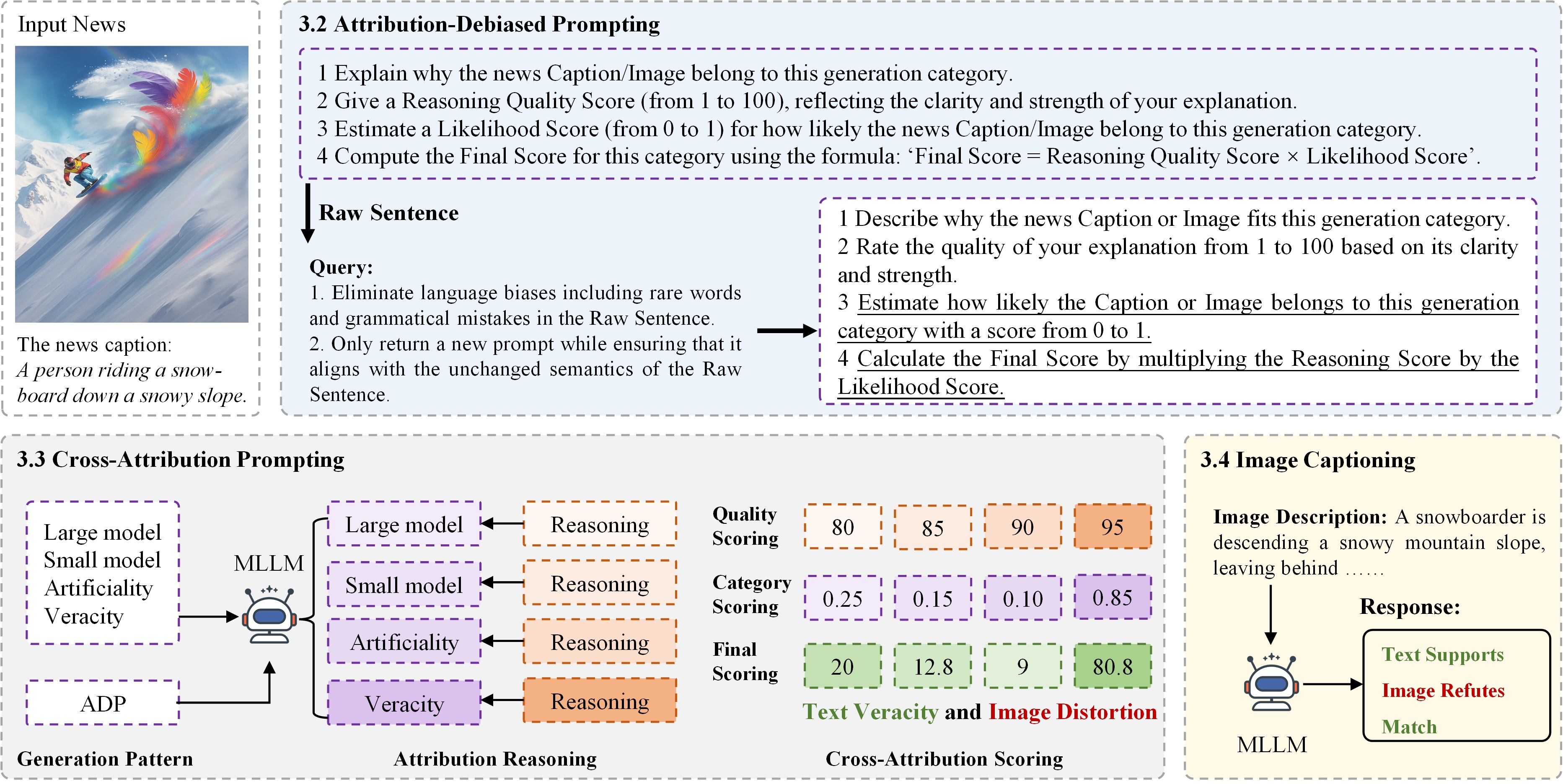}
\caption{The overall architecture of Insight-A, which consists of automatic attribution-debiased prompting, cross-attribution prompting, and image captioning.}
\label{fig2}
\end{figure*}

%% file: 4_experiment.tex
\section{Experiment}
\label{sec:Experiment}

\input{latex/tab_1}

\subsection{Experimental Setups}

\paragraph{Datasets} To evaluate the efficiency of our Insight-A, we conduct extensive experiments on the widely used benchmark dataset MMFakeBench \cite{liu2024mmfakebench}. This dataset comprises 11,000 image-text pairs, which intertwine mixed-source and multiple-type multimodal content. Moreover, it is split into validation and test sets with a ratio of 1:10.

\paragraph{Evaluation Metrics} To ensure fairness, we are consistent with previous studies \cite{chen_etal_2023_causal,beigi2025llmsimprovemultimodalfactchecking,liu2024mmfakebench}, employing the following evaluation metrics: F1, Pre, Rec, and ACC. Expressly, F1 is the widely adopted macro-F1 score, balancing precision and recall through their harmonic mean. Likewise, Pre, Rec, and ACC represent precision, recall, and accuracy as supplementary evaluation metrics.

\paragraph{Comparison Methods} To demonstrate the effectiveness and generalizability of Insight-A against SOTA methods in the setting of zero-shot, we select the following representative MLLMs as foundational models for comprehensive comparison: VILA \cite{Lin_2024_CVPR}, InstructBLIP \cite{NEURIPS2023_9a6a435e}, BLIP-2 \cite{pmlr_v202_li23q}, and LLaVA-1.6 \cite{liu2024llavanext}. These large models range from 13B parameters to 34B parameters. In MLLMs with 13B parameters, VILA, InstructBLIP, and BLIP-2 include LLaMA2 \cite{touvron2023llama}, Vicuna \cite{chiang2023vicuna}, and FlanT5-XXL \cite{JMLR_23_0870}. In addition, LLaVA-1.6 contains two types: Vicuna with 13B parameters, and Vicuna with 34B parameters.

Moreover, all experimental results are obtained by utilizing one NVIDIA A800 GPU (80GB) with PyTorch \cite{NEURIPS2019_bdbca288}. For consistency, we align hyper-parameters and setups with the SOTA MMD-Agent \cite{liu2024mmfakebench}.

\section{Overall Results}

In this part, we apply our Insight-A to the two baseline models, and illustrate the effectiveness and generalizability of Insight-A in the following settings: 1) multiclass evaluation, 2) binary evaluation, and 3) fine-grained evaluation.

\input{latex/tab_2}

\paragraph{Multiclass Evaluation.} In Table \ref{tab_1}, the experimental results emphasize the effectiveness and generalizability of Insight-A for multiclass classification in the realm of multimodal misinformation detection. Applied to LLaVA-1.6\textsubscript{Vicuna-34B}, Insight-A substantially boosts performance across all evaluation metrics, achieving the best results against the MMD-Agent. Typically, in the model LLaVA-1.6\textsubscript{Vicuna-13B}, Insight-A achieves 7.5\% and 6.7\% improvements over our baseline method in F1 score on the validation and test sets, respectively. In the model LLaVA-1.6\textsubscript{Vicuna-34B}, Insight-A still attains significant increases, underscoring our method's effectiveness. Moreover, Insight-A presents a consistent tendency on both validation and test sets, highlighting the Insight-A's strength in the generalization capabilities. In a nutshell, this strong performance across all evaluation metrics confirms the efficacy of Insight-A and its potential to advance multimodal misinformation detection capabilities.

\paragraph{Binary Evaluation.} From Table \ref{tab_2}, we can observe that our Insight-A achieves the highest F1 score, and outperforms other compared methods regarding binary overall results. In the 3-th row and the 6-th row of Table \ref{tab_2}, we provide the results when removing both MMD-Agent and Insight-A. The comparison between them shows that our Insight-A surpasses other methods by a large margin across different foundational models. Moreover, one can see that the MLLM with more parameters indeed introduces significant improvements in overall performance, which is consistent with the scaling law \cite{kaplan2020scalinglawsneurallanguage}. Along with the LLaVA-1.6\textsubscript{Vicuna-34B}, additional improvements of 1.1\% and 4.4\% are observed, showing the value of Insight-A even in larger models. Consistent results confirm the Insight-A's distinguishing capabilities.

\paragraph{Fine-Grained Evaluation.} In Table \ref{tab_3}, we compare Insight-A with other methods across different forgery sources, including TVD, VVD, and CCD. From the results, we can conclude that the majority of open-source models perform well in certain scenarios, but experience performance degradation in others. However, our Insight-A effectively reduces the performance discrepancies between different scenarios by a huge margin. Compared to the MMD-Agent with LLaVA-1.6\textsubscript{Vicuna-34B} across different forgery scenarios, our proposed method provides gains of 6\%, 21.3\%, 6.9\%, and 2.7\%, respectively. The results across different forgery sources effectively illustrate Insight-A's robustness in complex environments.

\input{latex/tab_3}
\input{latex/tab_4}

In Table \ref{tab_4}, we also provide a detailed comparison between Insight-A and MMD-Agent within the proprietary model (e.g., GPT-4V \cite{2023GPT4VisionSC}). Insight-A achieves the best performance on both multiclass and binary classification, reaching the new SOTA. Typically, although GPT-4V with standard prompting reaches a relatively high performance (see 1-th row in Table \ref{tab_4}), the lack of inherent reasoning prevents it from debunking erroneous information in the sense. Each method can enhance the model to understand and capture synthetic multimodal misinformation (see 2-th row to 4-th row in Table \ref{tab_4}), thus achieving a gain in performance. Furthermore, MMD-Agent with GPT-4V presents a high precision and a low recall, which indicates that it tends to make decisions for misinformation with identifiable points of falsification. However, Insight-A presents a good balance between precision and recall, highlighting its strength in debunking misinformation with both ambiguous contents and identifiable points of falsification.

To sum up, the experiments above confirm the effectiveness and generalizability of our Insight-A to the task of multimodal misinformation detection in a zero-shot setting.

\input{latex/tab_5}
\input{latex/tab_6}
\input{latex/tab_7}

\input{latex/tab_8}
\input{latex/tab_9}

\section{Ablation Study}


\paragraph{Impact of Models on ADP.} In Table \ref{tab_5}, we present the results of ADP by adopting different large models. LLMs include LLaMA-3 \cite{llama3modelcard}, Qwen-3 \cite{qwen3}, and DeepSeek-V3 \cite{deepseekai2024deepseekv3technicalreport}. MLLMs include GPT-4o \cite{hurst2024gpt}, Gemma-3 \cite{gemma_2025}, and LLaVA-1.6 \cite{liu2024llavanext}. From the results, we can achieve the following findings: The expert with domain knowledge performs the task adaptation prompting, resulting in accurate detections. The comparison between LLMs and MLLMs shows that MLLMs typically possess superior performance, which is attributed to their fine-tuning across multimodal tasks and datasets. On the other hand, the ADP on LLaVA-1.6 is consistent with the foundational model in Insight-A, thus leading to a significant gain in performance. Hence, leveraging the inherent capability of large models to the ADP can effectively enhance the detection performance of Insight-A. This promotes us to explore multimodal misinformation attribution from the perspective of MLLMs insight.

\paragraph{CAP and IC.} As shown in Table \ref{tab_6}, we further evaluate the effectiveness of CAP and IC in Insight-A with LLaVA-1.6\textsubscript{Vicuna-34B}. In 1-th row, we remove both CAP and IC components to achieve the results of our baseline method. The 2-th row and 3-th row refer to CAP and IC, respectively. Contrasting 1-th row with 2-th row verifies that deriving visual details via IC can incorporate informative cues into the final decision-making, thereby yielding performance gains. The comparison between 1-th row and 3-th row reveals that utilizing the CAP in our baseline method provides a large improvement in performance. In 4-th row, working with all modules further enhances the detection capabilities. The results clearly demonstrate the significance of both CAP and IC in our Insight-A.

\input{latex/fig_3}

\paragraph{Effect of Attribution.} To investigate the significance of multimodal misinformation attribution, we conduct extensive experiments as follows:
\begin{itemize}
    \item In Table \ref{tab_7}, we report the results on classification and attribution. Here, we leverage LLaVA-1.6 with 34B parameters as the foundational model. The F1 score is used as an evaluation metric for estimating attribution performance. The 1-th row denotes the results of our baseline method. Integrating attribution into our Insight-A significantly improves its performance (see 2-th row and 3-th row in Table \ref{tab_7}). Furthermore, we illustrate that our Insight-A results using 70\% of the ground truth attribution lead to the best performance on both multiclass and binary classification. This is expected because the ground truth attribution is achieved by decomposing the essence of misinformation production. The results confirm that predicting accurate attributions can ensure the high performance for multimodal misinformation detection.
    \item As shown in Figure \ref{fig3}, we exhibit the detection results of different methods on two prevalent generation categories. In text, the clear definition of AI-generated contents enables models to capture their boundary. Non AI-generated texts include compound generation categories, leading to a more substantial challenge. In image, synthetic images are created on real-life scenarios and AIGC technologies, limiting the perceptual capabilities of models for detecting them. Another type of image is edited by humans, which typically contains explicit forgery traces. This facilitates models to identify them. In both texts and images, our Insight-A leverages the inherent capabilities of MLLMs for reasoning answers, resulting in significant gains in performance. This underscores Insight-A's superiority in the task of multimodal misinformation detection.
    \item Finally, in Table \ref{tab_8}, we perform experiments for the effect of cross-attribution scoring in the CAP. From the results, we can note that each scoring way guides the model to catch forgery traces within multimodal contents, resulting in performance gains of 2.9\% and 3.4\%, respectively. This further confirms the significance of multimodal misinformation attribution. On the other hand, the mutual synergy between scoring ways (see 4-th row in Table \ref{tab_8}) brings a 4.7\% improvement, highlighting the strength of multiple reasoning paths in tasks demanding both perception and reasoning.
\end{itemize}

\input{latex/fig_4}

In summary, each designed module in Insight-A can promote it to effectively and accurately catch forgery sources and attributions, thereby leading to superior performance in the realm of multimodal misinformation detection.

\section{Further Analysis}

\paragraph{Analysis on Efficiency.} In Table \ref{tab_9}, we compare the inference time and GPU memory usage of different methods for measuring the efficiency of our Insight-A. From the results, we can observe that while Insight-A shows a slight increase in the inference time due to multiple attribution reasoning paths, its performance surpasses the SOTA by a significant margin. In addition, our Insight-A utilizes fewer computational resources, and provides higher performance. This confirms the efficiency of Insight-A in the deployment stage.

\paragraph{Analysis on Cases.} As shown in Figure \ref{fig4}, we present a challenging instance for analyzing the ability of Insight-A. Specifically, in text, Insight-A provides strong confidence levels of both perception and reasoning for a specific generation category, ensuring the news caption attribution. In image, Insight-A also provides high confidence levels for each attribution reasoning. However, the category scoring presents significant discrepancies between different generation categories. Therefore, such independent cross-attribution scoring effectively guides Insight-A to make correct decisions.

%% file: latex/tab_1.tex
\begin{table*}[t]
\centering
\setlength{\tabcolsep}{4.4mm}
\resizebox{\textwidth}{!}{
\begin{tabular}{llcccccccc}
\toprule
\multirow{2}{*}{\textbf{Method}} & \multirow{2}{*}{\textbf{Model}} & 
\multicolumn{4}{c}{\textbf{Validation (1000)}} & 
\multicolumn{4}{c}{\textbf{Test (10000)}} \\
\cmidrule(lr){3-6} \cmidrule(lr){7-10}
& & \textbf{F1$\uparrow$} & \textbf{Pre$\uparrow$} & \textbf{Rec$\uparrow$} & \textbf{ACC$\uparrow$} & \textbf{F1$\uparrow$} & \textbf{Pre$\uparrow$} & \textbf{Rec$\uparrow$} & \textbf{ACC$\uparrow$} \\
\midrule
\multirow{5}{*}{MMD-Agent} 
& VILA\textsubscript{LLaMA2-13B} & 22.7 & 27.3 & 24.4 & 28.7 & 24.0 & 30.4 & 25.5 & 29.4 \\
& InstructBLIP\textsubscript{Vicuna-13B} & 26.0 & 33.3 & 30.1 & 29.5 & 24.5 & 32.1 & 28.8 & 27.3 \\
& BLIP2\textsubscript{FlanT5-XXL} & 31.6 & 39.8 & 32.2 & 34.4 & 28.8 & 39.0 & 30.4 & 32.1 \\
& LLaVA-1.6\textsubscript{Vicuna-13B} & 38.0 & 44.5 & 41.0 & 40.6 & 34.5 & 42.7 & 37.5 & 37.4 \\
& LLaVA-1.6\textsubscript{Vicuna-34B} & 49.9 & 54.4 & 52.9 & 48.7 & 47.7 & 52.1 & 49.6 & 46.6 \\
\midrule
\multirow{2}{*}{Insight\textsubscript{Baseline}}
& LLaVA-1.6\textsubscript{Vicuna-13B}     & 34.6 & 43.1 & 38.9 & 40.1 & 34.5 & 43.6 & 38.2 & 39.5 \\
& LLaVA-1.6\textsubscript{Vicuna-34B}     & 53.7 & 53.7 & 56.4 & 54.9 & 53.2 & 52.3 & 55.5 & 55.0  \\
\midrule
\multirow{4}{*}{\textbf{Insight-A}} 
& LLaVA-1.6\textsubscript{Vicuna-13B}     & 42.1 & 50.9 & 50.0 & 43.0 & 41.2 & 51.0 & 48.5 & 41.9 \\
& \textit{Imp.(\%)}                       & +\textit{7.5} & +\textit{7.8} & +\textit{11.1} & +\textit{2.9} & +\textit{6.7} & +\textit{7.4} & +\textit{10.3} & +\textit{2.4} \\
& LLaVA-1.6\textsubscript{Vicuna-34B}     & \textbf{59.1} & \textbf{60.9} & \textbf{59.2} & \textbf{57.0} & \textbf{58.6} & \textbf{62.0} & \textbf{57.6} & \textbf{56.1} \\
& \textit{Imp.(\%)}                       & +\textit{5.4} & +\textit{7.2} & +\textit{2.8} & +\textit{2.1}  &  +\textit{5.4} & +\textit{9.7} & +\textit{2.1} & +\textit{1.1}   \\
\bottomrule
\end{tabular}
}
\caption{Multiclass results (\%) of different detection methods with different MLLMs on the benchmark dataset MMFakeBench. Best results are in bold.}
\label{tab_1}
\end{table*}

%% file: latex/tab_2.tex
\begin{table}[t]
\centering
\resizebox{\columnwidth}{!}{
\begin{tabular}{lcccccc}
\toprule
\textbf{Model} & \textbf{LLM} & \textbf{Size} & \textbf{MMD-Agent} & \textbf{Insight-A} & \textbf{Validation} & \textbf{Test} \\
\midrule
LLaVA-1.6 & Vicuna & 13B & \ding{51} & \ding{55} & 51.8 & 50.2 \\
LLaVA-1.6 & Vicuna & 13B & \ding{55} & \ding{51} & \textbf{56.1} & \textbf{57.7} \\
LLaVA-1.6 & Vicuna & 13B & \ding{55} & \ding{55} & 50.6 & 52.5 \\
\midrule
LLaVA-1.6 & Vicuna & 34B & \ding{51} & \ding{55} & 67.2 & 68.1 \\
LLaVA-1.6 & Vicuna & 34B & \ding{55} & \ding{51} & \textbf{68.3} & \textbf{72.5} \\
LLaVA-1.6 & Vicuna & 34B & \ding{55} & \ding{55} & 67.6 & 70.2 \\
\bottomrule
\end{tabular}
}
\caption{Binary overall results (F1 score (\%)) of different detection methods under MLLMs on both validation and test sets of MMFakeBench. Best results are in bold.}
\label{tab_2}
\end{table}

%% file: latex/tab_3.tex
\begin{table}[t]
\centering
\resizebox{\columnwidth}{!}{
\begin{tabular}{llccccc}
\toprule
\textbf{Method} & \textbf{Model} & \textbf{Real$\uparrow$} & \textbf{TVD$\uparrow$} & \textbf{VVD$\uparrow$} & \textbf{CCD$\uparrow$} & \textbf{Overall$\uparrow$} \\
\midrule
\multirow{5}{*}{MMD-Agent}
& VILA\textsubscript{LLaMA2-13B}            & 32.4 & 13.4 & 4.3  & 37.6 & 21.9 \\
& InstructBLIP\textsubscript{Vicuna-13B}    & 41.9 & 18.8 & 19.6 & 23.8 & 26.0 \\
& BLIP2\textsubscript{FlanT5-XXL}           & 41.5 & 39.2 & 13.1 & 32.6 & 31.6 \\
& LLaVA-1.6\textsubscript{Vicuna-13B}       & 17.9 & 51.5 & 30.9 & 42.3 & 35.7 \\
& LLaVA-1.6\textsubscript{Vicuna-34B}       & 51.1 & 37.6 & 61.7 & 49.2 & 49.9 \\
\midrule
\multirow{2}{*}{Insight\textsubscript{Baseline}}
& LLaVA-1.6\textsubscript{Vicuna-13B}       & 19.4 & 58.2 & 20.6 & 40.3 & 34.6 \\
& LLaVA-1.6\textsubscript{Vicuna-34B}       & 55.1 & \textbf{65.3} & 47.4 & 47.0 & 53.7 \\
\midrule
\multirow{2}{*}{\textbf{Insight-A}}
& LLaVA-1.6\textsubscript{Vicuna-13B}       & 42.4 & 45.4 & 36.4 & 44.0 & 42.1 \\
& LLaVA-1.6\textsubscript{Vicuna-34B}       & \textbf{57.1} & 58.9 & \textbf{68.6} & \textbf{51.9} & \textbf{59.1} \\
\bottomrule
\end{tabular}
}
\caption{Performance (F1 score (\%)) of different methods on different forgery sources of multimodal misinformation. Best results are in bold.}
\label{tab_3}
\end{table}

%% file: latex/tab_4.tex
\begin{table}[t]
\centering
\resizebox{\columnwidth}{!}{
\begin{tabular}{llcccccccc}
\toprule
\textbf{Category} & 
\multicolumn{4}{c}{\textbf{Multiclass Cls.}} & 
\multicolumn{4}{c}{\textbf{Binary Cls.}} \\
\cmidrule(lr){1-1} \cmidrule(lr){2-5} \cmidrule(lr){6-9}
\textbf{Method} & \textbf{F1$\uparrow$} & \textbf{Pre$\uparrow$} & \textbf{Rec$\uparrow$} & \textbf{ACC$\uparrow$} & \textbf{F1$\uparrow$} & \textbf{Pre$\uparrow$} & \textbf{Rec$\uparrow$} & \textbf{ACC$\uparrow$} \\
\midrule
GPT-4V\textsubscript{ChatGPT}               & 51.0 & 66.8 & 49.7 & 54.0 & 72.3 & 72.1 & 72.8 & 75.6 \\
\ + MMD-Agent                                 & 61.6 & \textbf{67.8} & 59.3 & 62.1 & 74.0 & 73.4 &	75.5 & 76.8 \\
\ + Insight\textsubscript{Baseline}           & 57.3 & 58.3 & 56.8 & 65.6 & 81.6 & 82.1 & 81.3 & 83.3 \\
\ + \textbf{Insight-A}                        & \textbf{61.8} & 61.5 & \textbf{62.3} & \textbf{67.5} & \textbf{82.2} & \textbf{82.3} & \textbf{82.1} &	\textbf{83.7} \\
\bottomrule
\end{tabular}
}
\caption{Performance (\%) comparison of different methods with the proprietary MLLM on both binary and multiclass classification (\textbf{Cls.}). Best results are in bold.}
\label{tab_4}
\end{table}

%% file: latex/tab_5.tex
\begin{table}[t]
\centering
\resizebox{\columnwidth}{!}{
\begin{tabular}{llccccc}
\toprule
\textbf{Category} & \textbf{Model} & \textbf{Real$\uparrow$} & \textbf{TVD$\uparrow$} & \textbf{VVD$\uparrow$} & \textbf{CCD$\uparrow$} & \textbf{Overall$\uparrow$} \\
\midrule
\multicolumn{2}{c}{Human}   & 56.7 & 58.6 &	63.5 & 51.3 & 57.5 \\
\midrule
\multirow{3}{*}{Language}
& LLaMA-3           & 43.5 & 41.5 & 56.8 & 44.5 & 46.6 \\
& Qwen-3            & 49.2 & 58.4 &	58.4 & 44.7 & 52.7 \\
& DeepSeek-V3       & 54.8 & 44.8 &	67.0 & 51.7 & 54.6 \\
\midrule
\multirow{3}{*}{Multimodal}
& GPT-4o            & 55.4 &	57.9 & 67.3 & 50.5 & 57.8 \\
& Gemma-3           & 55.9 &	57.4 & 68.3 & 51.5 & 58.3 \\
& LLaVA-1.6         & \textbf{57.1} & \textbf{58.9} & \textbf{68.6} & \textbf{51.9} & \textbf{59.1} \\
\bottomrule
\end{tabular}
}
\caption{Performance (F1 score (\%)) on different large models for the ADP. Best results are in bold.}
\label{tab_5}
\end{table}

%% file: latex/tab_6.tex
\begin{table}[t]
\centering
\resizebox{\columnwidth}{!}{
\begin{tabular}{lcccccc}
\toprule
\textbf{Model} & \textbf{+ CAP} & \textbf{+ IC} & \textbf{F1$\uparrow$} & \textbf{Pre$\uparrow$} & \textbf{Rec$\uparrow$} & \textbf{ACC$\uparrow$} \\
\midrule
\multirow{4}{*}{LLaVA-1.6}
& \ding{55}   &  \ding{55}      & 53.7 & 53.7 & 56.4 & 54.9 \\
& \ding{55}     &   \ding{51}    & 54.4 ($\uparrow$0.7) & 53.6 ($\downarrow$0.1) & 57.1 ($\uparrow$0.7) & 55.7 ($\uparrow$0.8) \\
& \ding{51} &   \ding{55}   & 58.7 ($\uparrow$5.0)  & 60.5 ($\uparrow$6.8) & 58.7 ($\uparrow$2.3) & 56.4 ($\uparrow$1.5) \\
& \ding{51} &  \ding{51}    & 59.1 ($\uparrow$5.4)  &	60.9 ($\uparrow$7.2) & 59.2 ($\uparrow$2.8) & 57.0 ($\uparrow$2.1) \\
\bottomrule
\end{tabular}
}
\caption{Results (\%) of ablations for the effect of CAP and IC on multiclass classification.}
\label{tab_6}
\end{table}

%% file: latex/tab_7.tex
\begin{table}[t]
\centering
\resizebox{\columnwidth}{!}{
\begin{tabular}{lcccccc}
\toprule
\textbf{Category} & \multicolumn{4}{c}{\textbf{Multiclass Cls.}} & \textbf{Binary Cls.} & \textbf{Attribution} \\
\cmidrule(lr){1-1} \cmidrule(lr){2-5} \cmidrule(lr){6-6} \cmidrule(lr){7-7}
\textbf{Model} & \textbf{F1$\uparrow$} & \textbf{Pre$\uparrow$} & \textbf{Rec$\uparrow$} & \textbf{ACC$\uparrow$} & \textbf{F1$\uparrow$} & \textbf{F1$\uparrow$}  \\
\midrule
\multirow{5}{*}{LLaVA-1.6}
& 53.7 & 53.7 &	56.4 & 54.9 & 67.6 & 42.1 \\
\rowcolor{lightgreen}
\multicolumn{7}{c}{Insight-A with predicted attributions} \\
& 59.1 & 60.9 &	59.2 & 57.0 & 68.3 & 52.8 \\
\rowcolor{lightgreen}
\multicolumn{7}{c}{Insight-A  with ground truth attributions} \\
& 68.6 & 67.3 & 71.2 & 67.9 & 74.0  & - \\
\bottomrule
\end{tabular}
}
\caption{Performance (\%) comparison on both classification (\textbf{Cls.}) and attribution.}
\label{tab_7}
\end{table}

%% file: latex/tab_8.tex
\begin{table}[t]
\centering
\setlength{\tabcolsep}{3.4mm}
\resizebox{\columnwidth}{!}{
\begin{tabular}{lccccc}
\toprule
\textbf{Model} & \textbf{LLM} & \textbf{Size} & \textbf{+ ARS} & \textbf{+ PPS} & \textbf{F1$\uparrow$} \\
\midrule
LLaVA-1.6 & Vicuna & 34B & \ding{55} & \ding{55} & 54.4 \\
LLaVA-1.6 & Vicuna & 34B & \ding{51} & \ding{55} & 57.3 ($\uparrow$2.9) \\
LLaVA-1.6 & Vicuna & 34B & \ding{55} & \ding{51} & 57.8 ($\uparrow$3.4) \\
LLaVA-1.6 & Vicuna & 34B & \ding{51} & \ding{51} & 59.1 ($\uparrow$4.7) \\
\bottomrule
\end{tabular}
}
\caption{Results (\%) of ablations for cross-attribution scoring in Insight-A. ARS denotes attribution reasoning scoring, and PPS denotes generation pattern scoring.}
\label{tab_8}
\end{table}

%% file: latex/tab_9.tex
\begin{table}[t]
\centering
\resizebox{\columnwidth}{!}{
\begin{tabular}{lccc}
\toprule
\textbf{Metric} & \textbf{MMD-Agent} & \textbf{Insight\textsubscript{Baseline}} &  \textbf{Insight-A} \\
\midrule
Average Inference Time (s) & 49.04s & 45.14s & 50.13s \\
GPU Memory Usage (GB) & 82G & 72G & 72G \\
Performance (F1) & 49.9 & 53.7 & 59.1 \\
\bottomrule
\end{tabular}
}
\caption{Efficiency comparison of different methods on the LLaVA-1.6 model with 34B parameters.}
\label{tab_9}
\end{table}

%% file: latex/fig_3.tex
\begin{figure}[t]
\centering
\includegraphics[width=0.7\columnwidth]{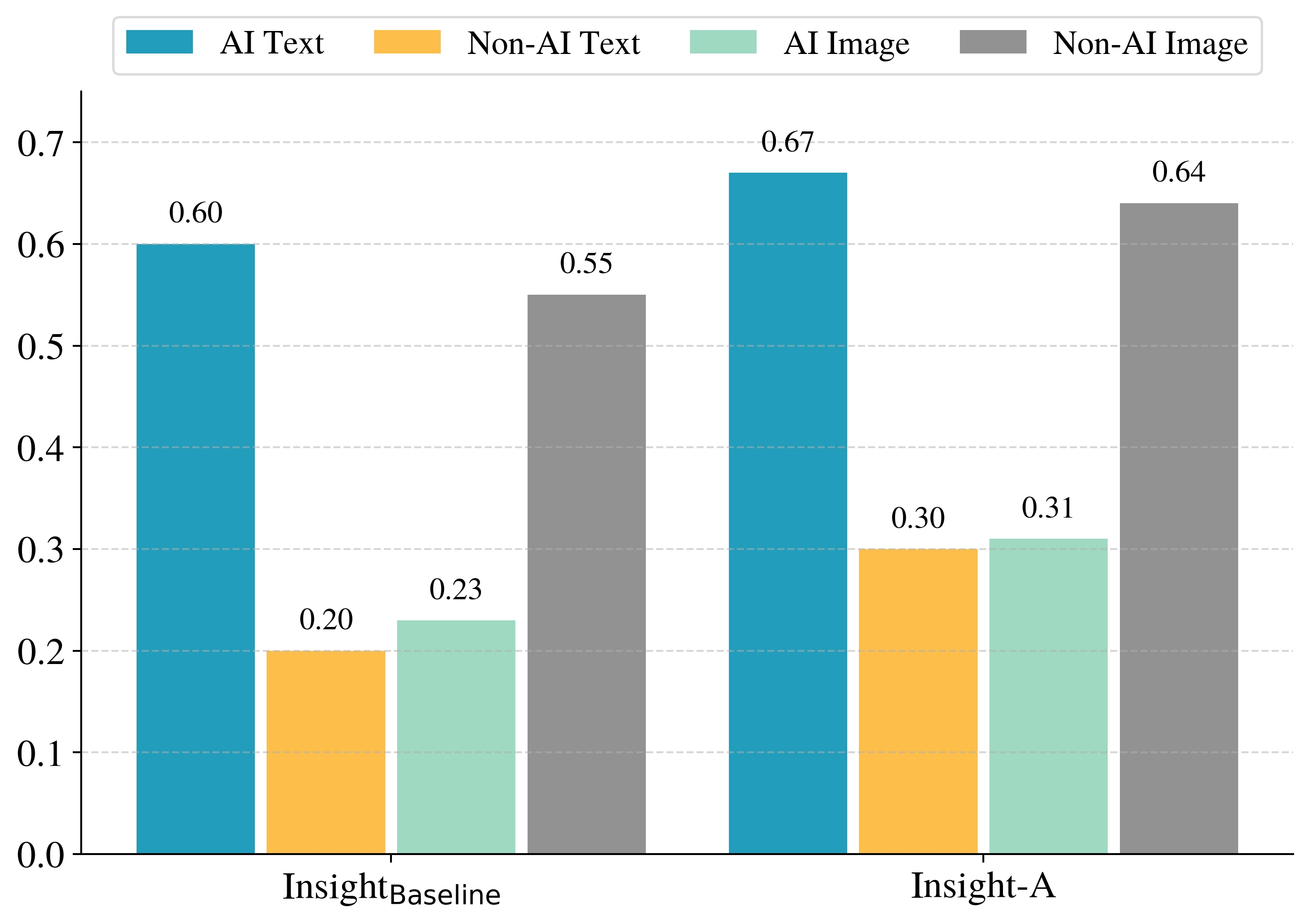}
\caption{Performance (detection success rate) of different methods on two generation categories.}
\label{fig3}
\end{figure}

%% file: latex/fig_4.tex
\begin{figure}[t]
\centering
\includegraphics[width=\columnwidth]{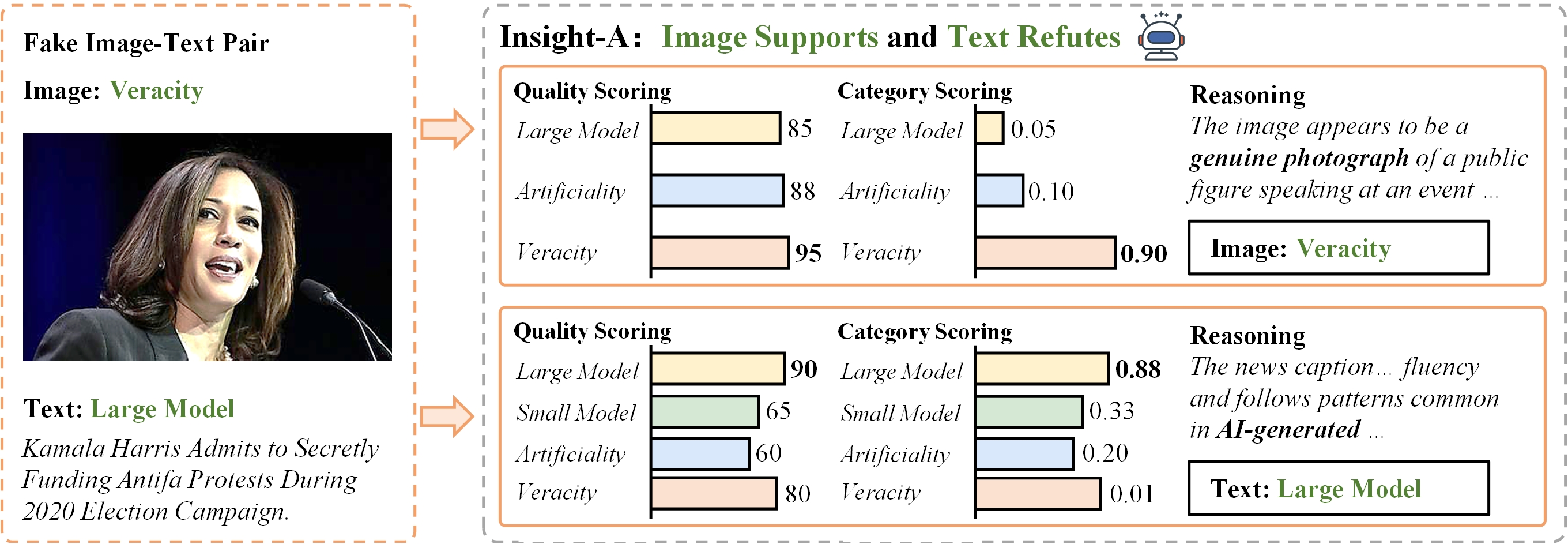}
\caption{The qualitative results of Insight-A.}
\label{fig4}
\end{figure}

%% file: 5_conclusion.tex
\section{Conclusion}
\label{sec:Conclusion}

This paper introduces Insight-A, a novel framework that explores attribution with MLLM insights for debunking misinformation in the era of AIGC. We emphasize the significance of multimodal misinformation attribution using multiple reasoning paths in Insight-A, underscoring it as a pivotal factor for detecting veracity distortion. Experiments across various evaluation settings demonstrate the effectiveness and generalizability of Insight-A, paving the way for combating deceptive misinformation created by AIGC technology nowadays.

%% file: main.bib
@inproceedings{liu2024mmfakebench,
  title={MMFakeBench: A Mixed-Source Multimodal Misinformation Detection Benchmark for LVLMs},
  author={Liu, Xuannan and Li, Zekun and Li, Peipei and Huang, Huaibo and Xia, Shuhan and Cui, Xing and Huang, Linzhi and Deng, Weihong and He, Zhaofeng},
  booktitle={ICLR},
  year={2025}
}

@article{wang2024mfcbenchbenchmarkingmultimodalfactchecking,
  title={Mfc-bench: Benchmarking multimodal fact-checking with large vision-language models},
  author={Wang, Shengkang and Lin, Hongzhan and Luo, Ziyang and Ye, Zhen and Chen, Guang and Ma, Jing},
  journal={arXiv preprint arXiv:2406.11288},
  year={2024}
}

@ARTICLE{10440475,
  author={Shao, Rui and Wu, Tianxing and Wu, Jianlong and Nie, Liqiang and Liu, Ziwei},
  journal={IEEE Transactions on Pattern Analysis and Machine Intelligence}, 
  title={Detecting and Grounding Multi-Modal Media Manipulation and Beyond}, 
  year={2024},
  volume={46},
  number={8},
  pages={5556-5574}
}

@inproceedings{NIPS2014_f033ed80,
 author = {Goodfellow, Ian J. and Pouget-Abadie, Jean and Mirza, Mehdi and Xu, Bing and Warde-Farley, David and Ozair, Sherjil and Courville, Aaron and Bengio, Yoshua},
 booktitle = {Advances in Neural Information Processing Systems},
 editor = {Z. Ghahramani and M. Welling and C. Cortes and N. Lawrence and K.Q. Weinberger},
 pages = {},
 publisher = {Curran Associates, Inc.},
 title = {Generative Adversarial Nets},
 volume = {27},
 year = {2014}
}

@InProceedings{Wang_2025_CVPR,
    author    = {Wang, Jin and Lv, Chenghui and Li, Xian and Dong, Shichao and Li, Huadong and Yao, Kelu and Li, Chao and Shao, Wenqi and Luo, Ping},
    title     = {Forensics-Bench: A Comprehensive Forgery Detection Benchmark Suite for Large Vision Language Models},
    booktitle = {Proceedings of the Computer Vision and Pattern Recognition Conference (CVPR)},
    month     = {June},
    year      = {2025},
    pages     = {4233-4245}
}

@article{Guo_Ma_Zeng_Luo_Zeng_Tang_Zhao_2025, title={Each Fake News Is Fake in Its Own Way: An Attribution Multi-Granularity Benchmark for Multimodal Fake News Detection}, volume={39}, DOI={10.1609/aaai.v39i1.31999}, abstractNote={Social platforms, while facilitating access to information, have also become saturated with a plethora of fake news, resulting in negative consequences. Automatic multimodal fake news detection is a worthwhile pursuit. Existing multimodal fake news datasets only provide binary labels of real or fake. However, real news is alike, while each fake news is fake in its own way. These datasets fail to reflect the mixed nature of various types of multimodal fake news. To bridge the gap, we construct an attributing multi-granularity multimodal fake news detection dataset AMG, revealing the inherent fake pattern. Furthermore, we propose a multi-granularity clue alignment model MGCA to achieve multimodal fake news detection and attribution. Experimental results demonstrate that AMG is a challenging dataset, and its attribution setting opens up new avenues for future research.}, number={1}, journal={Proceedings of the AAAI Conference on Artificial Intelligence}, author={Guo, Hao and Ma, Zihan and Zeng, Zhi and Luo, Minnan and Zeng, Weixin and Tang, Jiuyang and Zhao, Xiang}, year={2025}, month={Apr.}, pages={228-236} }

@article{Zhang_Zhang_Zhou_Huang_Li_2024, title={Reinforced Adaptive Knowledge Learning for Multimodal Fake News Detection}, volume={38}, abstractNote={Nowadays, detecting multimodal fake news has emerged as a foremost concern since the widespread dissemination of fake news may incur adverse societal impact. Conventional methods generally focus on capturing the linguistic and visual semantics within the multimodal content, which fall short in effectively distinguishing the heightened level of meticulous fabrications. Recently, external knowledge is introduced to provide valuable background facts as complementary to facilitate news detection. Nevertheless, existing knowledge-enhanced endeavors directly incorporate all knowledge contexts through static entity embeddings, resulting in the potential noisy and content-irrelevant knowledge. Moreover, the integration of knowledge entities makes it intractable to model the sophisticated correlations between multimodal semantics and knowledge entities. In light of these limitations, we propose a novel Adaptive Knowledge-Aware Fake News Detection model, dubbed AKA-Fake. For each news, AKA-Fake learns a compact knowledge subgraph under a reinforcement learning paradigm, which consists of a subset of entities and contextual neighbors in the knowledge graph, restoring the most informative knowledge facts. A novel heterogeneous graph learning module is further proposed to capture the reliable cross-modality correlations via topology refinement and modality-attentive pooling. Our proposal is extensively evaluated over three popular datasets, and experimental results demonstrate the superiority of AKA-Fake.}, number={15}, journal={Proceedings of the AAAI Conference on Artificial Intelligence}, author={Zhang, Litian and Zhang, Xiaoming and Zhou, Ziyi and Huang, Feiran and Li, Chaozhuo}, year={2024}, month={Mar.}, pages={16777-16785} }

@inproceedings{huang_etal_2024_miragenews,
    title = "{M}i{RAG}e{N}ews: Multimodal Realistic {AI}-Generated News Detection",
    author = "Huang, Runsheng  and
      Dugan, Liam  and
      Yang, Yue  and
      Callison-Burch, Chris",
    editor = "Al-Onaizan, Yaser  and
      Bansal, Mohit  and
      Chen, Yun-Nung",
    booktitle = "Findings of the Association for Computational Linguistics: EMNLP 2024",
    month = nov,
    year = "2024",
    address = "Miami, Florida, USA",
    publisher = "Association for Computational Linguistics",
    pages = "16436--16448"
}

@article{beigi2025llmsimprovemultimodalfactchecking,
  title={Can LLMs Improve Multimodal Fact-Checking by Asking Relevant Questions?},
  author={Beigi, Alimohammad and Jiang, Bohan and Li, Dawei and Tan, Zhen and Shaeri, Pouya and Kumarage, Tharindu and Bhattacharjee, Amrita and Liu, Huan},
  journal={arXiv preprint arXiv:2410.04616},
  year={2024}
}

@inproceedings{Xu_MDAM3,
author = {Xu, Qingzheng and Du, Heming and \L{}ukasik, Szymon and Zhu, Tianqing and Wang, Sen and Yu, Xin},
title = {MDAM3: A Misinformation Detection and Analysis Framework for Multitype Multimodal Media},
year = {2025},
isbn = {9798400712746},
publisher = {Association for Computing Machinery},
address = {New York, NY, USA},
booktitle = {Proceedings of the ACM on Web Conference 2025},
pages = {5285–5296},
numpages = {12},
location = {Sydney NSW, Australia},
series = {WWW '25}
}

@inproceedings{Min_2022,
author = {Min, Erxue and Rong, Yu and Bian, Yatao and Xu, Tingyang and Zhao, Peilin and Huang, Junzhou and Ananiadou, Sophia},
title = {Divide-and-Conquer: Post-User Interaction Network for Fake News Detection on Social Media},
year = {2022},
isbn = {9781450390965},
publisher = {Association for Computing Machinery},
address = {New York, NY, USA},
booktitle = {Proceedings of the ACM Web Conference 2022},
pages = {1148–1158},
numpages = {11},
location = {Virtual Event, Lyon, France},
series = {WWW '22}
}

@inproceedings{zhang_gao_2023_towards,
    title = "Towards {LLM}-based Fact Verification on News Claims with a Hierarchical Step-by-Step Prompting Method",
    author = "Zhang, Xuan  and
      Gao, Wei",
    editor = "Park, Jong C.  and
      Arase, Yuki  and
      Hu, Baotian  and
      Lu, Wei  and
      Wijaya, Derry  and
      Purwarianti, Ayu  and
      Krisnadhi, Adila Alfa",
    booktitle = "Proceedings of the 13th International Joint Conference on Natural Language Processing and the 3rd Conference of the Asia-Pacific Chapter of the Association for Computational Linguistics (Volume 1: Long Papers)",
    month = nov,
    year = "2023",
    address = "Nusa Dua, Bali",
    publisher = "Association for Computational Linguistics",
    pages = "996--1011"
}

@InProceedings{Liu_dino,
author="Liu, Shilong
and Zeng, Zhaoyang
and Ren, Tianhe
and Li, Feng
and Zhang, Hao
and Yang, Jie
and Jiang, Qing
and Li, Chunyuan
and Yang, Jianwei
and Su, Hang
and Zhu, Jun
and Zhang, Lei",
editor="Leonardis, Ale{\v{s}}
and Ricci, Elisa
and Roth, Stefan
and Russakovsky, Olga
and Sattler, Torsten
and Varol, G{\"u}l",
title="Grounding DINO: Marrying DINO with Grounded Pre-training for Open-Set Object Detection",
booktitle="Computer Vision -- ECCV 2024",
year="2025",
publisher="Springer Nature Switzerland",
address="Cham",
pages="38--55",
isbn="978-3-031-72970-6"
}

@article{Ying_Hu_Zhou_Qian_Zeng_Ge_2023, title={Bootstrapping Multi-View Representations for Fake News Detection}, volume={37}, abstractNote={Previous researches on multimedia fake news detection include a series of complex feature extraction and fusion networks to gather useful information from the news. However, how cross-modal consistency relates to the fidelity of news and how features from different modalities affect the decision-making are still open questions. This paper presents a novel scheme of Bootstrapping Multi-view Representations (BMR) for fake news detection. Given a multi-modal news, we extract representations respectively from the views of the text, the image pattern and the image semantics. Improved Multi-gate Mixture-of-Expert networks (iMMoE) are proposed for feature refinement and fusion. Representations from each view are separately used to coarsely predict the fidelity of the whole news, and the multimodal representations are able to predict the cross-modal consistency. With the prediction scores, we reweigh each view of the representations and bootstrap them for fake news detection. Extensive experiments conducted on typical fake news detection datasets prove that BMR outperforms state-of-the-art schemes.}, number={4}, journal={Proceedings of the AAAI Conference on Artificial Intelligence}, author={Ying, Qichao and Hu, Xiaoxiao and Zhou, Yangming and Qian, Zhenxing and Zeng, Dan and Ge, Shiming}, year={2023}, month={Jun.}, pages={5384-5392} }

@inproceedings{chen_etal_2023_causal,
    title = "Causal Intervention and Counterfactual Reasoning for Multi-modal Fake News Detection",
    author = "Chen, Ziwei  and
      Hu, Linmei  and
      Li, Weixin  and
      Shao, Yingxia  and
      Nie, Liqiang",
    editor = "Rogers, Anna  and
      Boyd-Graber, Jordan  and
      Okazaki, Naoaki",
    booktitle = "Proceedings of the 61st Annual Meeting of the Association for Computational Linguistics (Volume 1: Long Papers)",
    month = jul,
    year = "2023",
    address = "Toronto, Canada",
    publisher = "Association for Computational Linguistics",
    pages = "627--638"
}

@article{Yifang_Moment_GPT, title={Zero-shot Video Moment Retrieval via Off-the-shelf Multimodal Large Language Models}, volume={39}, url={https://ojs.aaai.org/index.php/AAAI/article/view/32971}, DOI={10.1609/aaai.v39i9.32971}, abstractNote={The target of video moment retrieval (VMR) is predicting temporal spans within a video that semantically match a given linguistic query. Existing VMR methods based on multimodal large language models (MLLMs) overly rely on expensive high-quality datasets and time-consuming fine-tuning. Although some recent studies introduce a zero-shot setting to avoid fine-tuning, they overlook inherent language bias in the query, leading to erroneous localization. To tackle the aforementioned challenges, this paper proposes Moment-GPT, a tuning-free pipeline for zero-shot VMR utilizing frozen MLLMs. Specifically, we first employ LLaMA-3 to correct and rephrase the query to mitigate language bias. Subsequently, we design a span generator combined with MiniGPT-v2 to produce candidate spans adaptively. Finally, to leverage the video comprehension capabilities of MLLMs, we apply Video-ChatGPT and span scorer to select the most appropriate spans. Our proposed method substantially outperforms the state-of-the-art MLLM-based and zero-shot models on several public datasets, including QVHighlights, ActivityNet-Captions, and Charades-STA.}, number={9}, journal={Proceedings of the AAAI Conference on Artificial Intelligence}, author={Xu, Yifang and Sun, Yunzhuo and Zhai, Benxiang and Li, Ming and Liang, Wenxin and Li, Yang and Du, Sidan}, year={2025}, month={Apr.}, pages={8978-8986} }

@InProceedings{Jiang_2024_CVPR,
    author    = {Jiang, Chaoya and Xu, Haiyang and Dong, Mengfan and Chen, Jiaxing and Ye, Wei and Yan, Ming and Ye, Qinghao and Zhang, Ji and Huang, Fei and Zhang, Shikun},
    title     = {Hallucination Augmented Contrastive Learning for Multimodal Large Language Model},
    booktitle = {Proceedings of the IEEE/CVF Conference on Computer Vision and Pattern Recognition (CVPR)},
    month     = {June},
    year      = {2024},
    pages     = {27036-27046}
}

@article{baichuan2023baichuan2,
  title={Baichuan 2: Open Large-scale Language Models},
  author={Baichuan},
  journal={arXiv preprint arXiv:2309.10305},
  year={2023}
}

@article{Restrepo_2025, title={Multi-OphthaLingua: A Multilingual Benchmark for Assessing and Debiasing LLM Ophthalmological QA in LMICs}, volume={39}, abstractNote={Current ophthalmology clinical workflows are plagued by over-referrals, long waits, and complex and heterogeneous medical records. Large language models (LLMs) present a promising solution to automate various procedures such as triaging, preliminary tests like visual acuity assessment, and report summaries. However, LLMs have demonstrated significantly varied performance across different languages in natural language question-answering tasks, potentially exacerbating healthcare disparities in Low and Middle-Income Countries (LMICs). This study introduces the first multilingual ophthalmological question-answering benchmark with manually curated questions parallel across languages, allowing for direct cross-lingual comparisons. Our evaluation of 6 popular LLMs across 7 different languages reveals substantial bias across different languages, highlighting risks for clinical deployment of LLMs in LMICs. Existing debiasing methods such as Translation Chain-of-Thought or Retrieval-augmented generation (RAG) by themselves fall short of closing this performance gap, often failing to improve performance across all languages and lacking specificity for the medical domain. To address this issue, We propose CLARA (Cross-Lingual Reflective Agentic system), a novel inference time de-biasing method leveraging retrieval augmented generation and self-verification. Our approach not only improves performance across all languages but also significantly reduces the multilingual bias gap, facilitating equitable LLM application across the globe.}, number={27}, journal={Proceedings of the AAAI Conference on Artificial Intelligence}, author={Restrepo, David and Wu, Chenwei and Tang, Zhengxu and Shuai, Zitao and Phan, Thao Nguyen Minh and Ding, Jun-En and Dao, Cong-Tinh and Gallifant, Jack and Dychiao, Robyn Gayle and Artiaga, Jose Carlo and Bando, André Hiroshi and Gracitelli, Carolina Pelegrini Barbosa and Ferrer, Vincenz and Celi, Leo Anthony and Bitterman, Danielle and Morley, Michael G and Nakayama, Luis Filipe}, year={2025}, month={Apr.}, pages={28321-28330} }

@misc{liu2024llavanext,
    title={LLaVA-NeXT: Improved reasoning, OCR, and world knowledge},
    author={Liu, Haotian and Li, Chunyuan and Li, Yuheng and Li, Bo and Zhang, Yuanhan and Shen, Sheng and Lee, Yong Jae},
    month={January},
    year={2024}
}

@InProceedings{Fang_2025_CVPR,
    author    = {Fang, Wenlong and Wu, Qiaofeng and Chen, Jing and Xue, Yun},
    title     = {Notes-guided MLLM Reasoning: Enhancing MLLM with Knowledge and Visual Notes for Visual Question Answering},
    booktitle = {Proceedings of the Computer Vision and Pattern Recognition Conference (CVPR)},
    month     = {June},
    year      = {2025},
    pages     = {19597-19607}
}

@article{Liu_Li_Rao_Gao_Guan_Li_Ma_2025, title={Union Is Strength! Unite the Power of LLMs and MLLMs for Chart Question Answering}, volume={39}, abstractNote={Chart Question Answering (CQA) requires models to perform chart perception and reasoning. Recent studies driven by Large Language Models (LLMs) have dominated CQA. These include employing more cognitively capable LLMs for indirectly reasoning over transformed charts, i.e., tables, and directly perceiving charts utilizing Multimodal Large Language Models (MLLMs) with a wider perceptual range. Yet, they often encounter bottlenecks due to the limitation of the receptive field of LLMs and the fragility of the complex reasoning of some MLLMs. To unite the strengths of LLMs and MLLMs to complement each other’s limitations, we propose Synergy, a framework that unites the power of both models for CQA. Synergy first unites the chart with a table as the augmented perceptual signal. Next, it unites LLMs and MLLMs, scheduling the former to decompose a question into subquestions and the latter to answer these by perceiving the chart. Lastly, it operates LLMs to summarize the subquestion-answer pairs to refine the final answer. Extensive experimental results on popular CharQA and PlotQA benchmarks reveal that, with the power of union, Synergy outperforms strong competitors and achieves superior boosts over naive MLLMs by uniting them with a smaller LLM.}, number={5}, journal={Proceedings of the AAAI Conference on Artificial Intelligence}, author={Liu, Jiapeng and Li, Liang and Rao, Shihao and Gao, Xiyan and Guan, Weixin and Li, Bing and Ma, Can}, year={2025}, month={Apr.}, pages={5487-5495} }

@InProceedings{Cai_2025_CVPR,
    author    = {Cai, Shengqu and Chan, Eric Ryan and Zhang, Yunzhi and Guibas, Leonidas and Wu, Jiajun and Wetzstein, Gordon},
    title     = {Diffusion Self-Distillation for Zero-Shot Customized Image Generation},
    booktitle = {Proceedings of the Computer Vision and Pattern Recognition Conference (CVPR)},
    month     = {June},
    year      = {2025},
    pages     = {18434-18443}
}

@article{Yao_Yuguang_Pan_Ning_Ye_Zhou_Xie_2025, title={StableVC: Style Controllable Zero-Shot Voice Conversion with Conditional Flow Matching}, volume={39}, abstractNote={Zero-shot voice conversion (VC) aims to transfer the timbre from the source speaker to an arbitrary unseen speaker while preserving the original linguistic content. Despite recent advancements in zero-shot VC using language model-based or diffusion-based approaches, several challenges remain: 1) current approaches primarily focus on adapting timbre from unseen speakers and are unable to transfer style and timbre to different unseen speakers independently; 2) these approaches often suffer from slower inference speeds due to the autoregressive modeling methods or the need for numerous sampling steps; 3) the quality and similarity of the converted samples are still not fully satisfactory. To address these challenges, we propose a Style controllable zero-shot VC approach named StableVC, which aims to transfer timbre and style from source speech to different unseen target speakers. Specifically, we decompose speech into linguistic content, timbre, and style, and then employ a conditional flow matching module to reconstruct the high-quality mel-spectrogram based on these decomposed features. To effectively capture timbre and style in a zero-shot manner, we introduce a novel dual attention mechanism with an adaptive gate, rather than using conventional feature concatenation. With this non-autoregressive design, StableVC can efficiently capture the intricate timbre and style from different unseen speakers and generate high-quality speech significantly faster than real-time. Experiments demonstrate that our proposed StableVC outperforms state-of-the-art baseline systems in zero-shot VC and achieves flexible control over timbre and style from different unseen speakers. Moreover, StableVC offers approximately 25x and 1.65x faster sampling compared to autoregressive and diffusion-based baselines.}, number={24}, journal={Proceedings of the AAAI Conference on Artificial Intelligence}, author={Yao, Jixun and Yuguang, Yang and Pan, Yu and Ning, Ziqian and Ye, Jianhao and Zhou, Hongbin and Xie, Lei}, year={2025}, month={Apr.}, pages={25669-25677} }

@article{ye2025generativepsycholexicalapproachconstructing,
  title={Generative psycho-lexical approach for constructing value systems in large language models},
  author={Ye, Haoran and Zhang, Tianze and Xie, Yuhang and Zhang, Liyuan and Ren, Yuanyi and Zhang, Xin and Song, Guojie},
  journal={arXiv preprint arXiv:2502.02444},
  year={2025}
}

@misc{barez_chain_2025,
  author = {Fazl Barez and Tung-Yu Wu and Iván Arcuschin and Michael Lan and Vincent Wang and Noah Siegel and Nicolas Collignon and Clement Neo and Isabelle Lee and Alasdair Paren and Adel Bibi and Robert Trager and Damiano Fornasiere and John Yan and Yanai Elazar and Yoshua Bengio},
  title = {Chain‑of‑Thought Is Not Explainability},
  publisher = {alphaXiv},
  year = {2025}
}

@InProceedings{Zhang_2024_CVPR,
    author    = {Zhang, Ji and Wu, Shihan and Gao, Lianli and Shen, Heng Tao and Song, Jingkuan},
    title     = {DePT: Decoupled Prompt Tuning},
    booktitle = {Proceedings of the IEEE/CVF Conference on Computer Vision and Pattern Recognition (CVPR)},
    month     = {June},
    year      = {2024},
    pages     = {12924-12933}
}

@inproceedings{li_attributed,
    title = "Improving Attributed Text Generation of Large Language Models via Preference Learning",
    author = "Li, Dongfang  and
      Sun, Zetian  and
      Hu, Baotian  and
      Liu, Zhenyu  and
      Hu, Xinshuo  and
      Liu, Xuebo  and
      Zhang, Min",
    editor = "Ku, Lun-Wei  and
      Martins, Andre  and
      Srikumar, Vivek",
    booktitle = "Findings of the Association for Computational Linguistics: ACL 2024",
    month = aug,
    year = "2024",
    address = "Bangkok, Thailand",
    publisher = "Association for Computational Linguistics",
    pages = "5079--5101"
}

@inproceedings{weinzierl_tree,
    title = "Tree-of-Counterfactual Prompting for Zero-Shot Stance Detection",
    author = "Weinzierl, Maxwell  and
      Harabagiu, Sanda",
    editor = "Ku, Lun-Wei  and
      Martins, Andre  and
      Srikumar, Vivek",
    booktitle = "Proceedings of the 62nd Annual Meeting of the Association for Computational Linguistics (Volume 1: Long Papers)",
    month = aug,
    year = "2024",
    address = "Bangkok, Thailand",
    publisher = "Association for Computational Linguistics",
    pages = "861--880"
}

@article{llama3modelcard,
  title={Llama 3 Model Card},
  author={AI@Meta},
  year={2024},
}

@article{qwen3,
    title={Qwen3 Technical Report}, 
    author={An Yang and Anfeng Li and Baosong Yang and Beichen Zhang and Binyuan Hui and Bo Zheng and Bowen Yu and Chang Gao and Chengen Huang and Chenxu Lv and Chujie Zheng and Dayiheng Liu and Fan Zhou and Fei Huang and Feng Hu and Hao Ge and Haoran Wei and Huan Lin and Jialong Tang and Jian Yang and Jianhong Tu and Jianwei Zhang and Jianxin Yang and Jiaxi Yang and Jing Zhou and Jingren Zhou and Junyang Lin and Kai Dang and Keqin Bao and Kexin Yang and Le Yu and Lianghao Deng and Mei Li and Mingfeng Xue and Mingze Li and Pei Zhang and Peng Wang and Qin Zhu and Rui Men and Ruize Gao and Shixuan Liu and Shuang Luo and Tianhao Li and Tianyi Tang and Wenbiao Yin and Xingzhang Ren and Xinyu Wang and Xinyu Zhang and Xuancheng Ren and Yang Fan and Yang Su and Yichang Zhang and Yinger Zhang and Yu Wan and Yuqiong Liu and Zekun Wang and Zeyu Cui and Zhenru Zhang and Zhipeng Zhou and Zihan Qiu},
    journal = {arXiv preprint arXiv:2505.09388},
    year={2025}
}

@article{deepseekai2024deepseekv3technicalreport,
  title={Deepseek-v3 technical report},
  author={Liu, Aixin and Feng, Bei and Xue, Bing and Wang, Bingxuan and Wu, Bochao and Lu, Chengda and Zhao, Chenggang and Deng, Chengqi and Zhang, Chenyu and Ruan, Chong and others},
  journal={arXiv preprint arXiv:2412.19437},
  year={2024}
}

@article{gemma_2025,
    title={Gemma 3},
    publisher={Kaggle},
    author={Gemma Team},
    year={2025}
}

@article{hurst2024gpt,
  title={Gpt-4o system card},
  author={Hurst, Aaron and Lerer, Adam and Goucher, Adam P and Perelman, Adam and Ramesh, Aditya and Clark, Aidan and Ostrow, AJ and Welihinda, Akila and Hayes, Alan and Radford, Alec and others},
  journal={arXiv preprint arXiv:2410.21276},
  year={2024}
}

@article{kaplan2020scalinglawsneurallanguage,
  title={Scaling laws for neural language models},
  author={Kaplan, Jared and McCandlish, Sam and Henighan, Tom and Brown, Tom B and Chess, Benjamin and Child, Rewon and Gray, Scott and Radford, Alec and Wu, Jeffrey and Amodei, Dario},
  journal={arXiv preprint arXiv:2001.08361},
  year={2020}
}

@inproceedings{wu_etal_2022_rap,
    title = "{R}a{P}: Redundancy-aware Video-language Pre-training for Text-Video Retrieval",
    author = "Wu, Xing  and
      Gao, Chaochen  and
      Lin, Zijia  and
      Wang, Zhongyuan  and
      Han, Jizhong  and
      Hu, Songlin",
    editor = "Goldberg, Yoav  and
      Kozareva, Zornitsa  and
      Zhang, Yue",
    booktitle = "Findings of the Association for Computational Linguistics: EMNLP 2022",
    month = dec,
    year = "2022",
    address = "Abu Dhabi, United Arab Emirates",
    publisher = "Association for Computational Linguistics",
    pages = "3036--3047"
}

@ARTICLE{7346469,
  author={Barrett, Daniel Paul and Barbu, Andrei and Siddharth, N. and Siskind, Jeffrey Mark},
  journal={IEEE Transactions on Pattern Analysis and Machine Intelligence}, 
  title={Saying What You're Looking For: Linguistics Meets Video Search}, 
  year={2016},
  volume={38},
  number={10},
  pages={2069-2081}
}

@InProceedings{Zhang_2025_CVPR,
    author    = {Zhang, Zhenxing and Wang, Yaxiong and Cheng, Lechao and Zhong, Zhun and Guo, Dan and Wang, Meng},
    title     = {ASAP: Advancing Semantic Alignment Promotes Multi-Modal Manipulation Detecting and Grounding},
    booktitle = {Proceedings of the Computer Vision and Pattern Recognition Conference (CVPR)},
    month     = {June},
    year      = {2025},
    pages     = {4005-4014}
}

@InProceedings{Lin_2024_CVPR,
    author    = {Lin, Ji and Yin, Hongxu and Ping, Wei and Molchanov, Pavlo and Shoeybi, Mohammad and Han, Song},
    title     = {VILA: On Pre-training for Visual Language Models},
    booktitle = {Proceedings of the IEEE/CVF Conference on Computer Vision and Pattern Recognition (CVPR)},
    month     = {June},
    year      = {2024},
    pages     = {26689-26699}
}

@inproceedings{NEURIPS2023_9a6a435e,
 author = {Dai, Wenliang and Li, Junnan and LI, DONGXU and Tiong, Anthony and Zhao, Junqi and Wang, Weisheng and Li, Boyang and Fung, Pascale N and Hoi, Steven},
 booktitle = {Advances in Neural Information Processing Systems},
 editor = {A. Oh and T. Naumann and A. Globerson and K. Saenko and M. Hardt and S. Levine},
 pages = {49250--49267},
 publisher = {Curran Associates, Inc.},
 title = {InstructBLIP: Towards General-purpose Vision-Language Models with Instruction Tuning},
 volume = {36},
 year = {2023}
}

@InProceedings{pmlr_v202_li23q,
  title = 	 {{BLIP}-2: Bootstrapping Language-Image Pre-training with Frozen Image Encoders and Large Language Models},
  author =       {Li, Junnan and Li, Dongxu and Savarese, Silvio and Hoi, Steven},
  booktitle = 	 {Proceedings of the 40th International Conference on Machine Learning},
  pages = 	 {19730--19742},
  year = 	 {2023},
  editor = 	 {Krause, Andreas and Brunskill, Emma and Cho, Kyunghyun and Engelhardt, Barbara and Sabato, Sivan and Scarlett, Jonathan},
  volume = 	 {202},
  series = 	 {Proceedings of Machine Learning Research},
  month = 	 {23--29 Jul},
  publisher =    {PMLR}
}

@inproceedings{NEURIPS2019_bdbca288,
 author = {Paszke, Adam and Gross, Sam and Massa, Francisco and Lerer, Adam and Bradbury, James and Chanan, Gregory and Killeen, Trevor and Lin, Zeming and Gimelshein, Natalia and Antiga, Luca and Desmaison, Alban and Kopf, Andreas and Yang, Edward and DeVito, Zachary and Raison, Martin and Tejani, Alykhan and Chilamkurthy, Sasank and Steiner, Benoit and Fang, Lu and Bai, Junjie and Chintala, Soumith},
 booktitle = {Advances in Neural Information Processing Systems},
 editor = {H. Wallach and H. Larochelle and A. Beygelzimer and F. d\textquotesingle Alch\'{e}-Buc and E. Fox and R. Garnett},
 pages = {},
 publisher = {Curran Associates, Inc.},
 title = {PyTorch: An Imperative Style, High-Performance Deep Learning Library},
 volume = {32},
 year = {2019}
}

@article{touvron2023llama,
  title={Llama 2: Open foundation and fine-tuned chat models},
  author={Touvron, Hugo and Martin, Louis and Stone, Kevin and Albert, Peter and Almahairi, Amjad and Babaei, Yasmine and Bashlykov, Nikolay and Batra, Soumya and Bhargava, Prajjwal and Bhosale, Shruti and others},
  journal={arXiv preprint arXiv:2307.09288},
  year={2023}
}

@article{chiang2023vicuna,
  title={Vicuna: An open-source chatbot impressing gpt-4 with 90\%* chatgpt quality},
  author={Chiang, Wei-Lin and Li, Zhuohan and Lin, Ziqing and Sheng, Ying and Wu, Zhanghao and Zhang, Hao and Zheng, Lianmin and Zhuang, Siyuan and Zhuang, Yonghao and Gonzalez, Joseph E and others},
  journal={See https://vicuna. lmsys. org (accessed 14 April 2023)},
  volume={2},
  number={3},
  pages={6},
  year={2023}
}

@article{JMLR_23_0870,
  author  = {Hyung Won Chung and Le Hou and Shayne Longpre and Barret Zoph and Yi Tay and William Fedus and Yunxuan Li and Xuezhi Wang and Mostafa Dehghani and Siddhartha Brahma and Albert Webson and Shixiang Shane Gu and Zhuyun Dai and Mirac Suzgun and Xinyun Chen and Aakanksha Chowdhery and Alex Castro-Ros and Marie Pellat and Kevin Robinson and Dasha Valter and Sharan Narang and Gaurav Mishra and Adams Yu and Vincent Zhao and Yanping Huang and Andrew Dai and Hongkun Yu and Slav Petrov and Ed H. Chi and Jeff Dean and Jacob Devlin and Adam Roberts and Denny Zhou and Quoc V. Le and Jason Wei},
  title   = {Scaling Instruction-Finetuned Language Models},
  journal = {Journal of Machine Learning Research},
  year    = {2024},
  volume  = {25},
  number  = {70},
  pages   = {1--53},
}

@inproceedings{2023GPT4VisionSC,
  title={GPT-4V(ision) System Card},
  author={OpenAI},
  year={2023},
}

@article{Qwen2VL,
  title={Qwen2-VL: Enhancing Vision-Language Model's Perception of the World at Any Resolution},
  author={Wang, Peng and Bai, Shuai and Tan, Sinan and Wang, Shijie and Fan, Zhihao and Bai, Jinze and Chen, Keqin and Liu, Xuejing and Wang, Jialin and Ge, Wenbin and Fan, Yang and Dang, Kai and Du, Mengfei and Ren, Xuancheng and Men, Rui and Liu, Dayiheng and Zhou, Chang and Zhou, Jingren and Lin, Junyang},
  journal={arXiv preprint arXiv:2409.12191},
  year={2024}
}

@inproceedings{ali-etal-2025-detection,
    title = "Detection of Human and Machine-Authored Fake News in {U}rdu",
    author = "Ali, Muhammad Zain  and
      Wang, Yuxia  and
      Pfahringer, Bernhard  and
      Smith, Tony C",
    editor = "Che, Wanxiang  and
      Nabende, Joyce  and
      Shutova, Ekaterina  and
      Pilehvar, Mohammad Taher",
    booktitle = "Proceedings of the 63rd Annual Meeting of the Association for Computational Linguistics (Volume 1: Long Papers)",
    month = jul,
    year = "2025",
    address = "Vienna, Austria",
    publisher = "Association for Computational Linguistics",
    pages = "3419--3428",
    ISBN = "979-8-89176-251-0"
}
